\PassOptionsToPackage{table,dvipsnames}{xcolor}
\documentclass[10pt,twocolumn,letterpaper]{article}

\usepackage{iccv}              % To produce the CAMERA-READY version
\usepackage{dsfont}
\usepackage{booktabs}
\usepackage{amsmath}
\usepackage{multirow}
\usepackage{graphicx}
\usepackage{xcolor}
\usepackage{colortbl}
\usepackage{fancybox}
\usepackage{multirow}
\usepackage{amssymb}% http://ctan.org/pkg/amssymb
\usepackage{pifont}% http://ctan.org/pkg/pifont
\newcommand{\cmark}{\ding{51}}%
\newcommand{\xmark}{\ding{55}}%
\usepackage{algorithm}
\usepackage{algpseudocode}
\usepackage[accsupp]{axessibility}  % Improves PDF readability for those with disabilities.

\definecolor{iccvblue}{rgb}{0.21,0.49,0.74}
\usepackage[pagebackref,breaklinks,colorlinks,allcolors=iccvblue]{hyperref}

\def\paperID{12606} % *** Enter the Paper ID here
\def\confName{ICCV}
\def\confYear{2025}

\title{FVGen: Accelerating Novel-View Synthesis with Adversarial Video Diffusion Distillation}

\author{
    Wenbin Teng$^{1, 2}$,
    Gonglin Chen$^{1, 2}$,
    Haiwei Chen$^{1, 2}$, 
    Yajie Zhao$^{1, 2}$\thanks{Corresponding author.}
    \and
    $^{1}$Institute for Creative Technologies
    \and
    $^{2}$University of Southern California
    \and
    {\tt\small \{wenbinte, gonglinc\}@usc.edu, \{chenh, zhao\}@ict.usc.edu},\\
}

\begin{document}
\vspace{-1cm}

\makeatletter
\let\@oldmaketitle\@maketitle% Store \@maketitle
\renewcommand{\@maketitle}{\@oldmaketitle% Update \@maketitle to insert...
  \centering
  \includegraphics[width=0.9\linewidth]
    {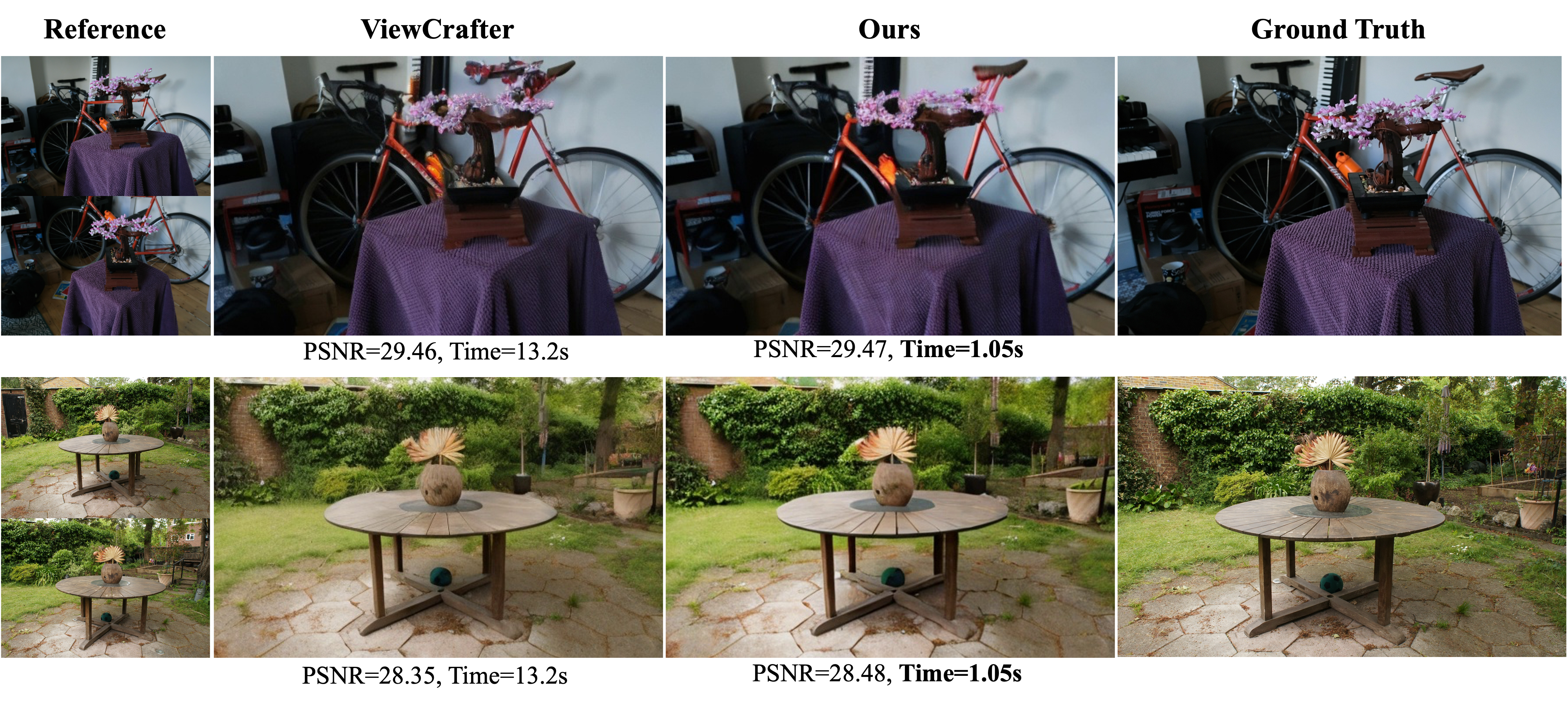}% ... an image
  \captionof{figure}{We present \textbf{FVGen}, a method of fast novel-view synthesis for 3D scene reconstruction with sparse inputs. Recent method~\cite{yu2024viewcrafter} unleash the generation capabilities of video diffusion models conditioned on a prior 3D cue for dense view creation, but always suffers from a long sampling time. We propose a novel framework that can significantly reduce the sampling time and maintain the generation quality.
  }
  \vspace{0 em}
}
\makeatother

\maketitle
\begin{abstract}

Recent progress in 3D reconstruction has enabled realistic 3D models from dense image captures, yet challenges persist with sparse views, often leading to artifacts in unseen areas. Recent works leverage Video Diffusion Models (VDMs) to generate dense observations, filling the gaps when only sparse views are available for 3D reconstruction tasks. A significant limitation of these methods is their slow sampling speed when using VDMs. In this paper, we present FVGen, a novel framework that addresses this challenge by enabling fast novel view synthesis using VDMs in as few as four sampling steps. We propose a novel video diffusion model distillation method that distills a multi-step denoising teacher model into a few-step denoising student model using Generative Adversarial Networks (GANs) and softened reverse KL-divergence minimization. Extensive experiments on real-world datasets show that, compared to previous works, our framework generates the same number of novel views with similar (or even better) visual quality while reducing sampling time by more than 90\%. FVGen significantly improves time efficiency for downstream reconstruction tasks, particularly when working with sparse input views (more than 2) where pre-trained VDMs need to be run multiple times to achieve better spatial coverage. Project Page: \href{https://wbteng9526.github.io/fvgen/}{https://wbteng9526.github.io/fvgen/}

\end{abstract}    
\vspace{-1em}
\section{Introduction}
\label{sec:intro}

Reconstructing 3D scenes from 2D images remains a pivotal research area at the intersection of computer vision and graphics. This field has extensive social impact, with broad applications in autonomous navigation, augmented reality, and more. With the recent development of neural rendering techniques such as NeRF~\cite{mildenhall2020nerf} and 3DGS~\cite{kerbl20233d}, creating high-quality 3D assets from 2D images has become more accessible. However, the success of 3D reconstruction depends on dense observations, which are often difficult to obtain in real-world scenarios. This limitation makes it challenging to apply these methods in many practical applications.

Introducing generative models into 3D reconstruction can fill gaps in missing views when only sparse input views are available. However, novel view synthesis from sparse views, especially in the settings with less than 2 views, is inherently an ill-posed problem. A bulk of previous works~\cite{yu2024viewcrafter, wu2024reconfusion, poole2022dreamfusion, chen2024v3d, gao2024cat3d} have found that priors learned from large image and video diffusion models~\cite{ho2020denoising, ho2022video} are rich enough to inpaint detailed and reasonable information in the many under-observed areas in the reconstructed 3D representations. We are primarily interested in the use of video diffusion models (VDM) to perform novel view synthesis. What bridges VDM to novel view synthesis is its ability to synthesize continuous views following a camera trajectories between the observed views. The spatio-temporal consistency in these smooth video trajectories, as demonstrated in several works~\cite{yu2024viewcrafter, chen2024v3d, xing2025dynamicrafter, chen2024mvsplat360, liu2024reconx}, effectively addresses the challenging "Janus Problem" that typically limits generation quality in other multi-view settings. However, VDM is a computationally expensive method because of the iterative sampling process that characterized all denoising diffusion methods. The long generation time makes VDM-based generation methods unsuitable for certain real-world downstream tasks, such as dynamic 3D reconstruction, where new consistent views must be generated continuously. This limitation also affects large-scale 3D reconstruction that requires synthesizing a significant number of novel views.

% To address the limitations, a line of recent methods, exemplified by ViewCrafter~\cite{yu2024viewcrafter}, leverages the generation capabilities of video diffusion model (VDM)~\cite{ho2022video} for photo-realistic novel views augmentation. To maintain 3D consistency, ViewCrafter conditions the generation of VDM with a 3D prior cue (i.e. DUSt3R). Through fine-tuning VDM with static 3D dataset~\cite{ling2024dl3dv,infinite_nature_2020}, it is able to generate both spatial-temporal consistent and 3D aware images, in significant support of downstream tasks like 3D reconstruction. 
% However, the performance of ViewCrafter is limited to its inference speed. 
% To generate photo-realistic frames, at least 50 steps are required for regular sampler like DDIM~\cite{song2020denoising}, which makes generating videos prohibitively slow. 
% Therefore, the long generation time makes method like ViewCrafter hard to deploy on real-world downstream tasks like dynamic 3D reconstruction with sparse inputs, where new consistent views are required to be generated constantly.

In this paper, we show that VDM-based novel view synthesis can be significantly accelerated for free. Without any compromise on the generation quality, we speed up diffusion-based synthesis of novel view by a factor of 90\% with a 4-step student model trained under a GAN objective~\cite{lin2025diffusion}. Acceleration of diffusion models has been actively studied in the past, but previous techniques may not be well suited to our task due to two reasons: First, we need a method specifically for accelerating video diffusion models, rather than image diffusion models. Second, the method must generalize effectively when trained on multi-view video datasets, which contain significantly less data than general video datasets. To accelerate a diffusion model, recent methods~\cite{yin2024one,sauer2024adversarial,meng2023distillation,lin2024sdxl,xu2024ufogen} propose minimizing the distributional difference between a few-step student generation model and the original multi-step teacher model. However, these methods are restricted to images with specific domains. The most related work to us is \cite{yin2024slow}, which generates more video frames with faster speed by optimizing a distribution matching distillation (DMD) loss. However, this method requires an effective initialization of the student model. This is completed by generating a noise-latent pair through an ODE solver~\cite{song2020denoising} and training the student model by optimizing a regression loss~\cite{yin2024one,yin2024slow}. The generation process usually takes a very long time and the student's ability is limited by the teacher's generation quality. More importantly, we have observed that DMD optimizations tend to be unstable and have the tendency towards mode collapse, when trained under multi-view video data. This may be due to the inherent concept of DMD loss optimization being the reverse KL-divergence minimization, which tends to be mode-seeking and zeroing out the modes of teacher's that are not relatively dominant.

% To accelerate diffusion model, recent methods~\cite{yin2024one} propose to minimize the difference in distribution between a few-step student generation model and the original multi-step teacher model. However, these methods are restricted in image dataset with specific domains. 

\begin{figure*}
    \centering
    \includegraphics[width=0.95\linewidth]{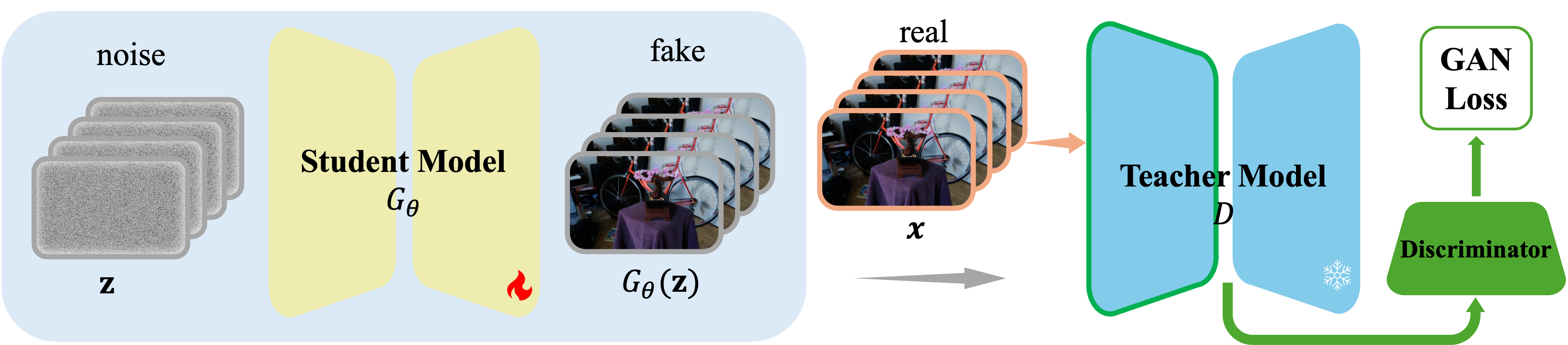}
    \caption{\textbf{Overview of student initialization.} We initialize our student model by training a GAN objective. The student model $G_\theta$, initialized with the weights of teacher model, uses few-step denoising to generate fake videos to fool the discriminator $D$. The fixed teacher model, together with a trainable discriminator, are optimized to differentiate between fake and real videos. }
    \label{fig:architecture_gan}
    \vspace{-1 em}
\end{figure*}
To solve the previous problems with VDM acceleration, we propose a novel framework, named \textbf{FVGen}, that performs novel views generation with a fast video diffusion generation with as few as 4 steps. First of all, different from previous GAN-based distillation methods~\cite{yin2025improved, sauer2024adversarial}, we find it beneficial to initial the few-step student model by training a GAN objective, where the student model is regarded as a generator of fake samples, and, instead of training a separate discriminator as in previous works, we leverage a pre-trained teacher model as the discriminator. Second, we propose the soften reverse KL-divergence to solve the unstable optimization of DMD loss. The soften reverse KL-divergence maintains similar mode-seeking behavior as reverse KL-divergence but with more robust distillation: it prevents the student from ignoring entire space of teacher's distribution, thereby preserving more of teacher's knowledge, which we find particularly beneficial in the context of limited data.

Through effective and robust distillation, FVGen is capable of generating video sequences with comparable or better visual quality as ViewCrafter~\cite{yu2024viewcrafter} but with significantly less time. FVGen is trained with Dl3DV-10K~\cite{ling2024dl3dv} dataset and extensive experiments on real world dataset including MipNeRF360~\cite{Barron2021Mip-NeRF:Fields} and Tanks-and-Temples~\cite{knapitsch2017tanks} demonstrate that our method achieves overall better performance than other state-of-the-art (SOTA) novel view synthesis methods and diffusion distillation methods.

\section{Related Works}
\label{sec:2_relatedworks}

\subsection{Diffusion-based Novel View Synthesis} \label{sec:2_3_relatedworks}

NeRF~\cite{mildenhall2020nerf} and 3DGS~\cite{kerbl20233d} excel at novel view synthesis with dense input views. However, reconstructing scenes from sparse views requires additional priors. Diffusion models~\cite{ho2020denoising,rombach2022high} can generate realistic pseudo input views from sparse original inputs. Several works use Score-Distillation Sampling~\cite{poole2022dreamfusion,wang2023score,lin2023magic3d,wang2024prolificdreamer} to create 3D objects from text prompts or single images by distilling latent diffusion model~\cite{rombach2022high} knowledge into 3D representations. While NeRF, 3DGS, and SDS typically require time-consuming per-scene optimization, view-dependent diffusion models~\cite{shi2023mvdream,gao2024cat3d,li2024director3d,kong2024eschernet} offer an alternative. These models directly generate 3D-consistent novel views conditioned on input images but directly lifting 2D images to 3D input therefore face challenges when scaling to larger scenes.

An alternative approach leverages video diffusion models~\cite{liu2024reconx,yu2024viewcrafter,chen2024v3d,voleti2025sv3d, chen2024mvsplat360} for multi-view image generation. These methods fine-tune pre-trained latent video diffusion models (LVDM)~\cite{blattmann2023align,xing2025dynamicrafter,hong2022cogvideo,yang2024cogvideox,chen2024videocrafter2} using multi-view images with both 2D and 3D guidance. The 2D guidance typically involves semantic features like CLIP~\cite{ilharco_gabriel_2021_5143773} embeddings, while 3D guidance comes from sparse input cues. For instance, \cite{yu2024viewcrafter} uses DUSt3R~\cite{wang2024dust3r} to build an initial point cloud and guides generation by combining point cloud renders with camera poses. Similarly, \cite{chen2024mvsplat360} conditions generation on feed-forward 3DGS renders, enabling 360-degree scene generation with longer frame sequences through multi-step refinement. However, these methods suffer from speed issues caused by video diffusion sampling when generating novel views.

\subsection{Diffusion Distillation}
Diffusion models typically require numerous denoising steps to generate high-quality samples, which makes them computationally intensive. To accelerate generation, distillation methods train a student model that mimics the teacher diffusion model's behavior while using fewer sampling steps. Using fewer sampling steps reduces quality, so existing works incorporate adversarial training~\cite{chen2024nitrofusion,kang2024distilling,luo2024you,sauer2024fast,xu2024ufogen,sauer2024adversarial,lin2024sdxl} to enhance the student model's output. However, simple GAN training cannot fully bridge the gap between student and teacher distributions, causing the student's generated output to remain noticeably different from the original diffusion model. To address this problem, Distribution Matching Distillation (DMD)~\cite{yin2024one,yin2025improved} optimizes the reverse KL-divergence. For tractable optimization, ~\cite{yin2024one,yin2025improved} leverages the gradient of DMD loss, which represents the difference between student and teacher score functions. For fast video generation, ~\cite{yin2024slow} extends DMD to video diffusion models like CogVideoX~\cite{yang2024cogvideox}. While these methods work well for general image and video generation tasks, they suffer from mode collapse and training instability when applied to view synthesis on relatively small datasets like multi-view video datasets. 
\section{Preliminaries}
\begin{figure*}
    \centering
    \includegraphics[width=0.95\linewidth]{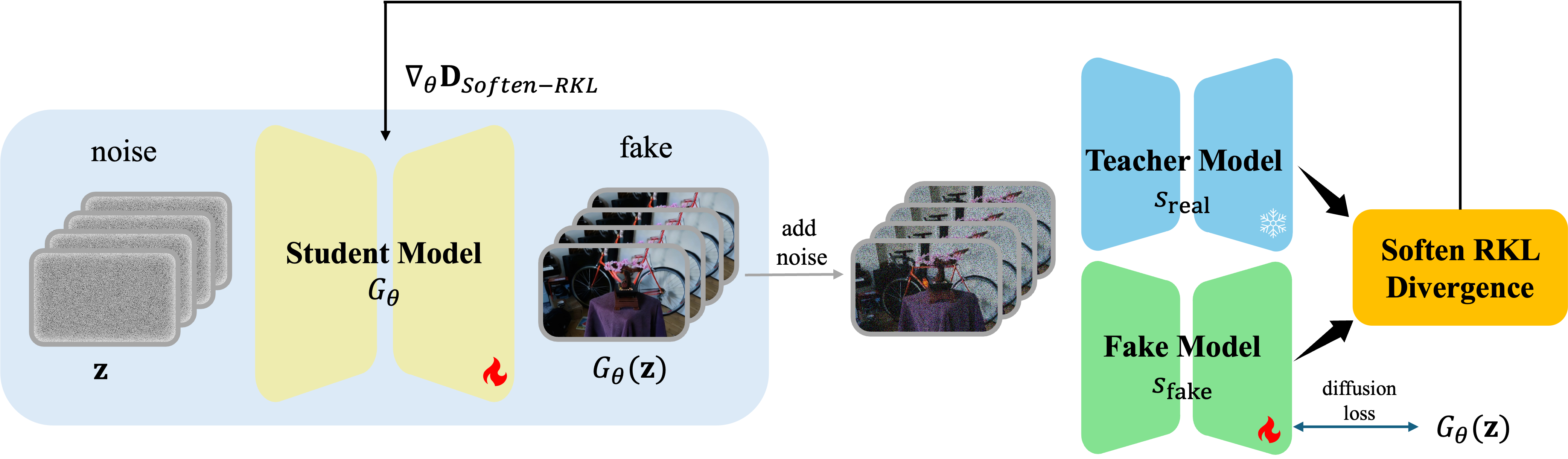}
    \caption{\textbf{Overview of Distribution Matching Distillation (DMD).} The gradient is the difference between the teacher score and fake score w.r.t. the student model output. We apply soften reverse KL-divergence for stable training and avoiding mode-collapse.}
    \label{fig:architecture_dmd}
    \vspace{-1em}
\end{figure*}
\subsection{Video Diffusion Model} \label{sec:3_1_prelim}
A video diffusion model is a generative model designed to synthesize realistic video sequences by learning the complex temporal and spatial patterns within video data. Based on the diffusion process, these models iteratively apply noise to video frames and then learn to reverse this process to generate new sequences. The forward diffusion process progressively adds Gaussian noise to the input clean sample $\boldsymbol{x}_0 \sim p(\boldsymbol{x})$ over $T$ timesteps, leading to highly noisy data $\boldsymbol{x}_T \sim \mathcal{N}(0, \boldsymbol{I})$. This process is formulated as:

\vspace{-1em}
\begin{equation} \label{eqn:3_prelim_vdmqsample}
    q(\boldsymbol{x}_t | \boldsymbol{x}_{t-1})=\mathcal{N}(\boldsymbol{x}_t;\sqrt{\alpha_t}\boldsymbol{x}_{t-1},(1-\alpha_t)\boldsymbol{I}),
\end{equation}

\noindent where $\alpha_t$ and $\boldsymbol{x}_t$ are the noise strength and noisy data at timestep $t$. The model then learns a denoising function, often parameterized by a neural network $\epsilon_\theta(\boldsymbol{x}_t,t)$, to reverse this process and generate realistic video frames, where the optimization function is defined by the MSE loss:
\begin{equation} \label{eqn:3_prelim_vdmloss}
    \mathcal{L}_{\text{diff}}(\theta)=\mathbb{E}_{\scriptstyle t\sim \mathcal{U}(0,1),\epsilon\sim\mathcal{N}(0,I),\boldsymbol{x}\sim p,c}\Big\| \epsilon - \epsilon_\theta(\boldsymbol{x}_t,t,c) \Big\|^2,
\end{equation}

\noindent where $c$ is the condition embedding of diffusion model which is usually represented as a text or image prompt. Instead of generating sequences at full resolution, Latent Video Diffusion Model (LVDM) first encodes video data $\boldsymbol{x}\in \mathbb{R}^{F\times C\times H\times W}$ into latent space using a pre-trained VAE~\cite{kingma2013auto} encoder: $\boldsymbol{z}=\mathcal{E}(\boldsymbol{x})$, where $\boldsymbol{z}\in \mathbb{R}^{F\times C \times h \times w}$. The compression will mitigate computational complexity while maintaining video generation quality. 

\subsection{Distribution Matching Distillation}
\label{sec:pre_dmd}
Distribution matching distillation adopts the idea of variational score distillation~\cite{wang2024prolificdreamer} that is optimized to match the distribution of student and teacher model by minimizing the reverse KL-divergence. Given a student model $G$ parameterized by $\theta$, the gradient of reverse KL-divergence is
\begin{align}
    \nabla\mathcal{L}_{\text{DMD}} &= \mathbb{E}_{t} \left( \nabla_{\theta} \textbf{D}_\text{Reverse-KL} (p_{\text{fake}, t} \parallel p_{\text{real}, t}) \right) \nonumber\\
% &= - \mathbb{E}_{t} \Bigg(p_{\text{fake}} \log\frac{p_\text{real}}{p_\text{fake}}\Bigg)\\
&= - \mathbb{E}_{t} \Bigg(  \bigg[ s_{\text{real}} ( F(G_{\theta}(\boldsymbol{z}), t), t) \nonumber \\
&- s_{\text{fake}} ( F(G_{\theta}(\boldsymbol{z}), t), t )) \frac{d G_{\theta}(\boldsymbol{z})}{d \theta} \bigg]\Bigg), 
\label{eqn:dmdloss}
\end{align}

\noindent where $p_{\text{real}}$ and $p_{\text{fake}}$ are the real and fake distribution. $\boldsymbol{z}\sim \mathcal{N}(0,\boldsymbol{I})$ is the initial random Gaussian noise. $F$ is the diffusion forward function determined by Eq.~(\ref{eqn:3_prelim_vdmqsample}). $s_{\text{real}}$ and $s_{\text{fake}}$ are the real and fake score function and defined as follows according to~\cite{song2020score}:
\vspace{-0.5em}
\begin{equation}
    s(\boldsymbol{x}_t, t) = \nabla_{\boldsymbol{x}_t} \log p_{t} (\boldsymbol{x}_t) = -\frac{\boldsymbol{x}_t - \alpha_t \mu(\boldsymbol{x}_t, t)}{\sigma_t^2},
\end{equation}

\noindent where $p_t$ is the distribution of the noisy sample $\boldsymbol{x}$. $\mu$ is the clean sample prediction of the diffusion model. While optimizing student model $G_\theta$ with gradient descent formulated in Eq.~(\ref{eqn:dmdloss}), the gradients of $\mu_\text{real}$ are stopped and $\mu_\text{fake}$ is dynamically optimized with the output of $G_\theta$ with regular diffusion loss similar to Eq.~(\ref{eqn:3_prelim_vdmloss})

\section{Method}

% Our method generates N views between the camera trajectory of two given views X1, X2 

We propose Fast Video Generation (\textbf{FVGen}) that distills the generation capability of a multi-step teacher 3D-VDM into a few-step student 3D-VDM. The teacher 3D-VDM used in our experiments is the novel-view synthesis model ViewCrafter~\cite{yu2024viewcrafter}. From a set of observed images $\boldsymbol{I}^{\text{obs}}$ and observed cameras $\boldsymbol\pi^{\text{obs}}$, ViewCrafter learns a conditional distribution $\boldsymbol{I}^{\text{tgt}} \sim p(\boldsymbol{I}^{\text{tgt}} \vert \boldsymbol{I}^{\text{obs}}, \boldsymbol\pi^{\text{obs}}, \boldsymbol\pi^{\text{tgt}})$, where $\boldsymbol{I}^{\text{tgt}} = \{ \boldsymbol{I}^\prime_i \}^L_{i=1}$ is a sequence of $L$ images corresponding to the target views $\boldsymbol\pi^{\text{tgt}}$. The student model FVGen is trained to sample from the same conditional distribution with a much faster inference speed. The architectures of FVGen are shown in Figure~\ref{fig:architecture_gan} and~\ref{fig:architecture_dmd}. For notation simplicity purpose, we ignore including $\boldsymbol{I}^{\text{obs}}, \boldsymbol\pi^{\text{obs}}, \boldsymbol\pi^{\text{tgt}}$ in the future formula as they are all the conditional inputs for both teacher and student diffusion model.

We first initialize the parameters of student model by training with a generative adversarial network (GAN) objective, leveraging the teacher model as discriminator (Section~\ref{sec:gan_training}). After that, we continue to train the student model by optimizing DMD loss and diffusion loss to minimize the soften reverse KL-divergence between the distribution of student and teacher model (Section~\ref{sec:f-divergence}).
\subsection{Student Initialization with GAN Training}
\label{sec:gan_training}
Directly training student model with DMD (Eq.~\ref{eqn:dmdloss}) will cause training collapse as it is difficult to naively distill multi-step denoising into few-step. \cite{yin2024one,yin2024slow} proposed to create noise-latent pairs by an ODE solver~\cite{song2020denoising} and minimize a regression loss with student model for initialization. However, we found that it is very time consuming to generate large amount of video data pairs and the student model will also be restricted by the teacher's limitations. Inspired by~\cite{yin2025improved, lin2025diffusion}, we propose to initialize the student model by training a GAN objective with the real images. Specifically, the student model ($G_\theta$) generates fake video samples that fool the teacher model who serves as a discriminator (denoted as $D$) by minimizing the generator loss $\mathcal{L}_G$. The discriminator $D$ classifies real samples from generated fake samples by maximizing $-\mathcal{L}_{D}$. The min-max game adversarial optimization is formulated as:
\vspace{-0.5em}
\begin{align}
\label{eq:gan_loss}
    \mathcal{L}_D &= \mathbb{E}_{\substack{\boldsymbol{x}\sim{p_\text{real}} \\ t \sim \mathcal{U}[0,T]}} \left[ \log f \left( D(F(\boldsymbol{x}, t)) \right) \right] \nonumber \\
&- \mathbb{E}_{\substack{\boldsymbol{z} \sim \mathcal{N}(0,\boldsymbol{I}) \\ t \sim \mathcal{U}[0,T]}} \left[ \log f \left( D(F(G_\theta(\boldsymbol{z}), t)) \right) \right], \\
\mathcal{L}_G &= \mathbb{E}_{\substack{\boldsymbol{z} \sim \mathcal{N}(0,\boldsymbol{I}) \\ t \sim \mathcal{U}[0,T]}} \left[ \log f \left( D(F(G_\theta(\boldsymbol{z}), t)) \right) \right], \nonumber
\end{align}

\noindent where $t$ is the time-step sampling and $F$ is the forward noise-injection model similar to Eq.~(\ref{eqn:dmdloss}). Discriminator $D$ retrieves the output feature map of the middle block of diffusion model UNet and passes it into the classifier $f$. To fit with video data, $f$ is composed of a 3D Convolutional Neural Network that compresses the feature map into classification logits. Here, we use the original sample $\boldsymbol{x}$ as the input to the discriminator instead of the generated sample $\boldsymbol{x}$ from teacher model such that the student model would not be restricted by the generation capability of teacher model.%Similar to StyleGAN2~\cite{Karras2020ada}, we apply softplus for classification output: $\texttt{Softplus}(x)=\frac{1}{\beta}\log(1+e^{\beta x})$. 
For the purpose of training stability, recent methods~\cite{yu2024viewcrafter,chen2024mvsplat360} predict the velocity field $\boldsymbol{v}_\theta$, and we convert it into the sample prediction with $\boldsymbol{x}_0=\sqrt{\alpha_t}\boldsymbol{x}_t-\sqrt{1-\alpha_t}\boldsymbol{v}_\theta$. Although the generated video samples are still very blurry, the GAN training scheme provides an effective initialization of the student model, which paves the way for more effective training with distribution matching loss.
\begin{figure*}
    \centering
    \includegraphics[width=\linewidth]{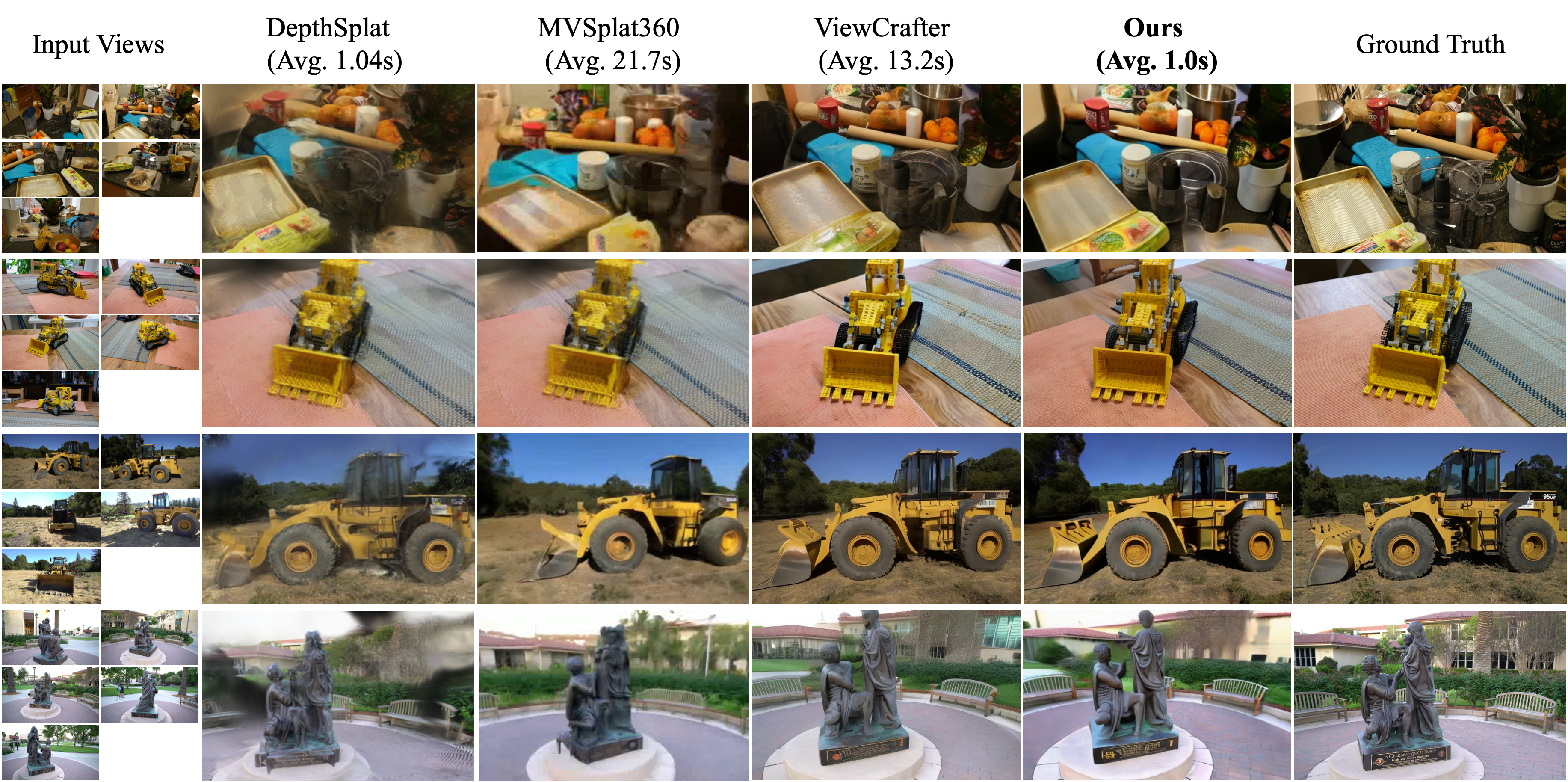}
    \caption{\textbf{Qualitative Results.} We compare FVGen with SOTA novel view synthesis method on MipNeRF360~\cite{Barron2021Mip-NeRF:Fields} and Tanks-and-Temples~\cite{knapitsch2017tanks} datasets. }
    \label{fig:qual_fig1}
    \vspace{-1 em}
\end{figure*}

\subsection{Distribution Matching with Soften Reverse KL-Divergence Minimization}
\label{sec:f-divergence}
As discussed in Section~\ref{sec:pre_dmd}, the core concept of distribution matching distillation (DMD)~\cite{yin2024one,yin2025improved} is to minimize the reverse KL-divergence between student and teacher distributions. While reverse KL is popular for its mode-seeking focus, it has notable limitations in the context of generative modeling, its tendency toward mode collapse. Since reverse KL rewards the student for zeroing out probability in any region where the teacher's density is low, the student can effectively ignore parts where the teacher's distribution are not dominant. If the teacher's distribution $p_\text{real}(\boldsymbol{x})$ has multiple modes of varying height, minimizing Reverse-KL$(p_\text{fake} \|p_\text{real})$ may lead $p_\text{fake}$ to concentrate on a subset of those modes and assign negligible mass to others. Therefore, the exclusive nature of reverse KL-divergence makes it prone to mode dropping and a lack of coverage of the full distribution.

To address the shortcomings of pure reverse KL, we propose to leverage a $Soften$ reverse KL-divergence (Soften RKL)~\cite{ghosh_kl_divergence,xu2025one}. Soften RKL refers to a modified divergence that retains the mode-seeking bias of reverse KL but softens its exclusion other relatively lower-probability regions, thereby mitigating mode collapse. Compared to reverse KL formulated in Eq.~(\ref{eqn:dmdloss}), the Soften RKL can be formulated as:
\vspace{-1em}
\begin{align}
    \mathbf{D}_\text{Soften-RKL}(p_{\text{fake},t}\| p_{\text{real},t})
    &=\mathbf{D}_{KL}\bigg(\frac{1}{2}p_\text{real}+\frac{1}{2}p_\text{fake}\bigg\|p_\text{real}\bigg), \nonumber \\
    &=p_\text{fake}(r+1)\log\bigg(\frac{1}{2}+\frac{1}{2r}\bigg),
\end{align}
% \vspace{-0.5em}
\noindent where $r=r(\boldsymbol{x},t):=p_{\text{real},t}(\boldsymbol{x})/p_{\text{fake},t}(\boldsymbol{x})$ is the density ratio. Here $\boldsymbol{x}$ denotes the noise-injection version of the output of student model formulated as $F(G_\theta(\boldsymbol{z}),t)$ with $t\sim\mathcal{U}(0,T)$. Thanks to our optimized GAN objective in Section~\ref{sec:gan_training}, the density ratio could be approximated by the following:
\begin{equation}
    r(\boldsymbol{x},t)=\frac{p_{\text{real},t}(\boldsymbol{x})}{p_{\text{fake},t}(\boldsymbol{x})} \approx \frac{f(D(\boldsymbol{x},t))}{1-f(D(\boldsymbol{x},t))}.
\end{equation}

\noindent In summary, the well-optimized GAN objective not only provides an efficient weight initialization of the student model, but also provides a direct estimate of the density ratio. Here, instead of directly comparing $p_\text{fake}$ and $p_\text{real}$ as in $\mathbf{D}_{KL}(p_\text{fake}\|p_\text{real})$, we compare an even mixture of $p_\text{real}$ and $p_\text{fake}$ to $p_\text{real}$. Soften RKL maintains the mode-seeking property of reverse KL, while penalizes more on mode dropping. This criterion yields a more robust distillation since it prevents the student from ignoring the entire teacher distribution, thereby preserving more of teacher's knowledge. Through experiments, we find that the soften RKL yields more stable training and more realistic generations. Computing the divergence directly is generally intractable, the gradient with respect to $\theta$ is formulated as follows:
\begin{align}
\label{eq:softenrkl_dmd}
    &\nabla\mathcal{L}_{\text{DMD}}^{\texttt{Soften-RKL}}= \\
    &- \mathbb{E}_{t} \Bigg( \frac{1}{r(\boldsymbol{x},t)}\bigg[ s_{\text{real}} ( \boldsymbol{x},t) - s_{\text{fake}} ( \boldsymbol{x},t)\bigg] \frac{d G_{\theta}(\boldsymbol{z})}{d \theta}\Bigg), \nonumber
\end{align}

\noindent where $\boldsymbol{x}=F(G_\theta(z),t)$ as previously defined. Apart from the soften RKL divergence optimization, we dynamically update the fake diffusion model $\mu_\text{fake}$ to adjust to change of student distribution in accordance with the original setting of DMD~\cite{yin2024one}. The loss term is identical to Eq.~(\ref{eqn:3_prelim_vdmloss}). We summarize the full training procedure in Alg.~\ref{alg:fsrvd_training}
\begin{algorithm}
    \caption{FVGen Training Procedure}
    \label{alg:fsrvd_training}
    \begin{algorithmic}[1]
    \Require Few-step timesteps $\mathcal{T} = \{0, t_1, t_2, \dots, t_Q\}$, pre-trained teacher model $\mu_\text{real}$, discriminator classifier $D$ dataset $\mathcal{D}$.
    \State \textbf{Initialize} student model $G_\theta$ with $\mu_\text{real}$.
    \State \textbf{Initialize} fake score function with $\mu_\text{fake}$.
    \While {training}
        \State Sample a video from dataset $\boldsymbol{x}_0\sim\mathcal{D}$
        \State Add noise with timestep $t\sim\mathcal{T}$: $\boldsymbol{x}_t=\sqrt{\alpha_t}\boldsymbol{x}_0+\sqrt{1-\alpha_t}\boldsymbol{\epsilon}, \quad \boldsymbol{\epsilon}\sim\mathcal{N}(0,\boldsymbol{I})$
        \State Predict with student: $\hat{\boldsymbol{x}}_0=G_\theta(\boldsymbol{x}_t,t)$
        \State Add noise $\boldsymbol\epsilon^\prime\sim\mathcal{N}(0,\boldsymbol{I})$ with timestep $\tau\sim\mathcal{U}(0,T)$ to real and fake sample:
        \State$\hat{\boldsymbol{x}}_\tau=\sqrt{\alpha_\tau}\hat{\boldsymbol{x}}_0+\sqrt{1-\alpha_\tau}\boldsymbol{\epsilon}^\prime$
        \State$\boldsymbol{x}_\tau=\sqrt{\alpha_\tau}\boldsymbol{x}_0+\sqrt{1-\alpha_\tau}\boldsymbol{\epsilon}^\prime$
        \State Update $G_\theta$ and $D$ with Eq.~(\ref{eq:gan_loss})
    \EndWhile
    \State \textbf{Output} trained student $G_\theta$
    \While {training}
    \State Repeat step 4 - 8
    \State Update $G_\theta$ with DMD Loss: Eq.~(\ref{eq:softenrkl_dmd})
    \State Add noise $\boldsymbol\epsilon^{\prime\prime}\sim\mathcal{N}(0,\boldsymbol{I})$ with $t_1\sim\mathcal{U}(0,T)$ to $\hat{\boldsymbol{x}}_0$:
    \State$\hat{\boldsymbol{x}}_{t_1}=\sqrt{\alpha_{t_1}}\hat{\boldsymbol{x}}_0+\sqrt{1-\alpha_{t_1}}\boldsymbol{\epsilon}^{\prime\prime}$
    \State Update $\mu_\text{fake}$ with denoising loss: Eq.~(\ref{eqn:3_prelim_vdmloss})
    \EndWhile
    \end{algorithmic}
\end{algorithm}
% \vspace{-2em}
\section{Experiments}
\subsection{Experiment Setup} \label{sec:5_1_exp}
\paragraph{Implementation Details.}

All of our student model, teacher model and fake score function is initialized with the parameters of the sparse model of ViewCrafter~\cite{yu2024viewcrafter}. We first initialize the student model by training GAN objective. The parameters of teacher model is fixed, and the student model and GAN classifier are trained for 4000 iterations with a two-scale update rule inspired by~\cite{yin2025improved}. Next, we optimize DMD loss by continuously training student model and the fake score function for another 5000 iterations with the same two-scale update rule. The whole pipeline is trained on 8 NVIDIA H100 with batch size 4. The training last about 1 day. We utilize AdamW optimizer~\cite{loshchilov2017decoupled} and set learning rates of all optimizers to be $5\times 10^{-5}$. The model is trained on images with a resolution of $512\times 320$.

\paragraph{Datasets.} FVGen is trained with DL3DV-10K~\cite{ling2024dl3dv}, a real-world, scene-level video dataset that includes more than 10K long video clips. Similar to~\cite{yu2024viewcrafter}, we construct point cloud and ground truth video pairs using DL3DV dataset. We randomly sample short video clips with a random frame stride less than 5. The first frame and last frame are used to construct the prior point cloud with DUSt3R~\cite{wang2024dust3r}. We use PyTorch3d~\cite{ravi2020pytorch3d}, an efficient renderer to construct the point cloud renders with the camera poses of intermediate frames. Through this process, we create 20,000 training data pairs. We validate our method on two public datasets Tanks-and-Temples~\cite{knapitsch2017tanks} and MipNeRF360~\cite{Barron2021Mip-NeRF:Fields}

\begin{figure*}
    \centering
    \includegraphics[width=\linewidth]{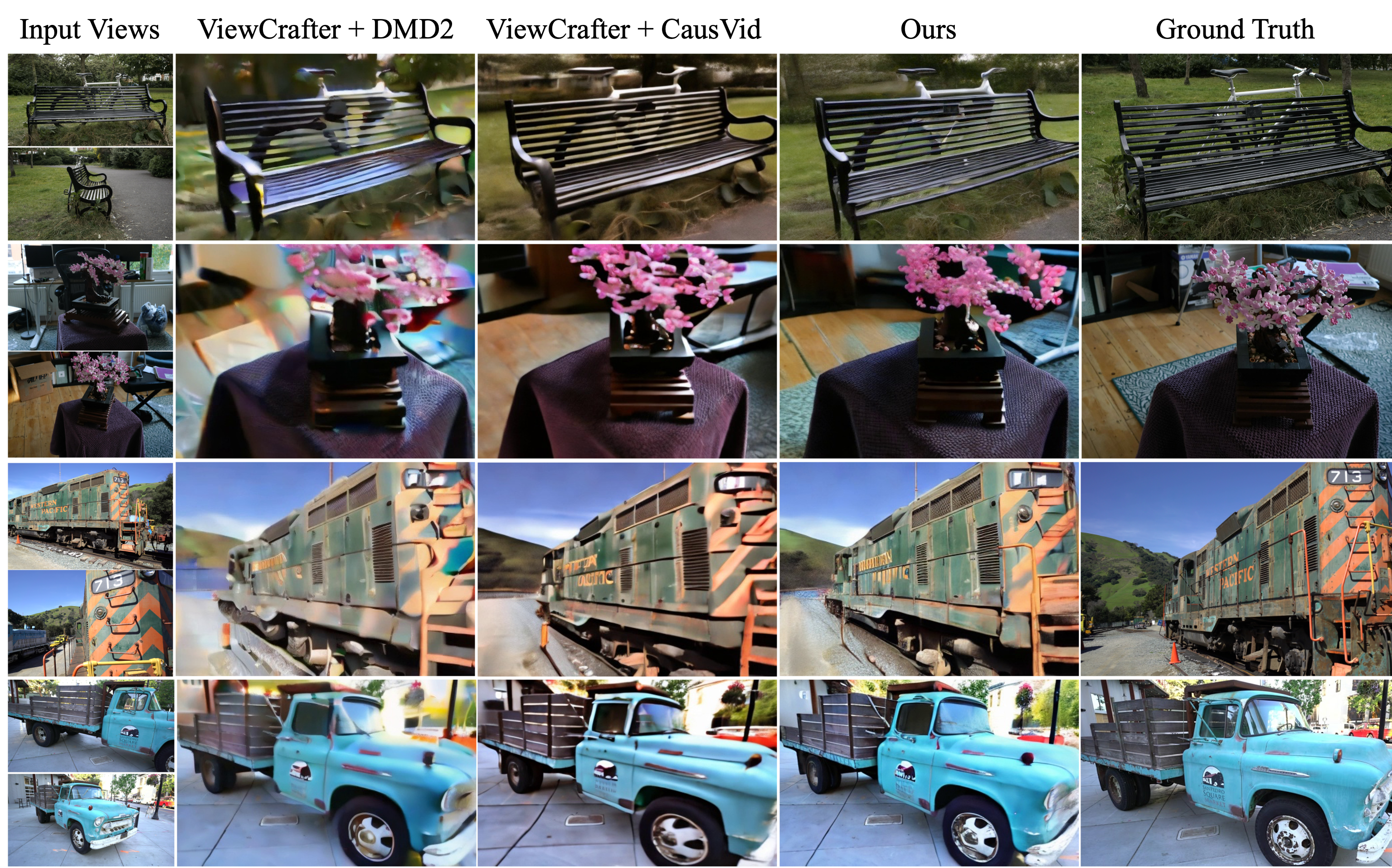}
    \caption{\textbf{Qualitative Results.} We compare FVGen with baseline diffusion distillation method on MipNeRF360~\cite{Barron2021Mip-NeRF:Fields} and Tanks-and-Temples~\cite{knapitsch2017tanks} datasets.  }
    \label{fig:qual_fig2}
    \vspace{-1em}
\end{figure*}

\paragraph{Baseline Methods and Evaluation Metrics.} 
%We compare the generation capability of our VDM with two regression-based novel view synthesis: MVSplat~\cite{chen2024mvsplat}, DepthSplat~\cite{xu2024depthsplat} and two diffusion based novel view synthesis method: ViewCrafter~\cite{yu2024viewcrafter} and DynamiCrafter~\cite{xing2025dynamicrafter}. 
To evaluate our proposed method, we compare our method with several state-of-the-art methods: ViewCrafter~\cite{yu2024viewcrafter}, MVSplat360~\cite{chen2024mvsplat360}. We also compare our video diffusion distillation method with two state-of-the-art methods: Caus-Vid~\cite{yin2024slow} and DMD2~\cite{yin2025improved}. We use PSNR~\cite{hore2010image}, SSIM~\cite{wang2004image}, LPIPS~\cite{zhang2018unreasonable} to measure reconstruction quality. In addition, we also reported the distribution metric, i.e. Frechet Inception Distance (FID), which compares the distribution of generated views and ground truth views. Apart from the visual quality comparison.

\subsection{Novel View Synthesis Comparison}
Similar to the evaluation setting of MVSplat360~\cite{chen2024mvsplat360}, for each scene of both MipNeRF360~\cite{Barron2021Mip-NeRF:Fields} and Tanks-and-Temples~\cite{knapitsch2017tanks}, we select 5 views that are far from each other but could cover the whole scene. We use our method and baseline method to generate 56 views sampled from the natural camera trajectory (14 views between each 2 input views).

\paragraph{Qualitative Results}
The qualitative comparisons with SOTA novel view synthesis methods are shown in Figure~\ref{fig:qual_fig1}. DepthSplat is a feed-forward generalizable Gaussian splatting method. When there are relatively low overlap between input frames, DepthSplat would exhibit obvious artifacts due to inconsistent depth scale and the issue of floating Gaussians. In comparison, MVSplat360~\cite{chen2024mvsplat360} uses VDM to refine the Gaussian splatting renders and remove the floaters. However, the current MVSplat360 only supports training with low resolution inputs and the current results present significant blurriness. Compared to DepthSplat and MVSplat360, ViewCrafter~\cite{yu2024viewcrafter} generates photo-realistic novel views with higher resolution. However, ViewCrafter is limited by the generation speed. Given generating 16 total frames, the average inference time is 13.2 seconds. In comparison, our method achieves a 10x speedup, generating the same number of frames in approximately 1 second and achieves similar visual results. The qualitative results prove that our method is able to distill the multi-step sampling into a few-step sampling process and achieve results that are similar to or even better than teacher ViewCrafter model with the student model.

\paragraph{Quantitative Analysis}
\begin{table}[h]
    \small
    \centering
    \renewcommand{\arraystretch}{1.0}
    \setlength{\tabcolsep}{2pt}
    \begin{tabular}{clccccc}
        \toprule
        % & \multicolumn{4}{c}{\textbf{ArtifactInjection}} \\
        % \cmidrule(lr){2-5} 
        & & PSNR $\uparrow$ & SSIM $\uparrow$ & LPIPS $\downarrow$ & FID $\downarrow$ & Time $\downarrow$ \\
        \midrule
        \multirow{4}{*}{\rotatebox[origin=c]{90}{\small{mip-360}}}
        &DepthSplat & 11.23 & 0.213 & 0.715 & 32.45 & \cellcolor{red!20}4.2\\
        &MVSplat360 & 12.28 & 0.285 & 0.682 & 25.69 & 87.2\\
        &ViewCrafter& \cellcolor{red!20}16.35 &  \cellcolor{yellow!40}0.346 & \cellcolor{yellow!40}0.433 & \cellcolor{red!20}16.28 & \cellcolor{yellow!40}66.3 \\
        
        &Ours & \cellcolor{yellow!40}16.28 & \cellcolor{red!20}0.352 & \cellcolor{red!20}0.429 & \cellcolor{yellow!40}17.44 & \cellcolor{yellow!40}5.1\\

        \midrule
        \multirow{4}{*}{\rotatebox[origin=c]{90}{\small{TNT}}}
        &DepthSplat & 12.43 & 0.263 & 0.677 & 35.88 & \cellcolor{red!20}4.3\\
        &MVSplat360 & 14.18 & 0.301 & 0.532 & 25.23 & 87.3\\
        &ViewCrafter & \cellcolor{yellow!40}18.69 & \cellcolor{yellow!40}0.402 & \cellcolor{red!20}0.208 & \cellcolor{yellow!40}23.94 & \cellcolor{yellow!40}65.9\\
        &Ours & \cellcolor{red!20}18.72 & \cellcolor{red!20}0.411 & \cellcolor{yellow!40}0.210 & \cellcolor{red!20}23.64 & \cellcolor{yellow!40}5.0\\

        \bottomrule
    \end{tabular}
    \caption{\textbf{Quantitative Results.} We report four quantitative metrics for image quality comparison on 2 separate datasets: MipNeRF360~\cite{Barron2021Mip-NeRF:Fields} (denoted as mip-360 in the table) and Tanks-and-Temples~\cite{knapitsch2017tanks} (denoted as TNT in the table). We highlight the best results in red and second-best in yellow.} 
    \label{tab:quant_nvs}
    \vspace{-2 em}
\end{table}
Table~\ref{tab:quant_nvs} presents quantitative comparisons on both datasets of FVGen and other novel view synthesis methods. Except for ViewCrafter~\cite{yu2024viewcrafter}, our method significantly surpasses other SOTA methods in terms of both perceptual and distribution metrics. Compared with ViewCrafter, our method yields similar quality but with more than 90\% faster in generation speed. The quantitative results also highlights the effectiveness of our proposed method.

\subsection{Diffusion Distillation Comparison}
We also compare FVGen with SOTA diffusion distillation methods. Similar to the 2-view evaluation setting of DepthSplat~\cite{xu2024depthsplat} and MVSplat~\cite{chen2024mvsplat}, we select 2 views that are far from each other in both MipNeRF360~\cite{Barron2021Mip-NeRF:Fields} and Tanks-and-Temples~\cite{knapitsch2017tanks} dataset. We use our method and baseline method to generate 16 views samples from the natural camera trajectory (14 intermediate views and 2 input views).

\paragraph{Qualitative Results}
First we perform DMD2~\cite{yin2025improved} on ViewCrafter for few-step distillation. DMD2 applies end-to-end training of student model, GAN discriminator and fake score function. We find that this training scheme is not stable and present significant variance so that it is hard to converge, leading to blurry results as visually depicted in Figure~\ref{fig:qual_fig2}. In addition, we compare our method with a recent work, Caus-Vid~\cite{yin2024slow}, on ViewCrafter distillation. Caus-Vid~\cite{yin2024slow} creates a small dataset with ODE solver and initialize the student model by optimizing a regression loss. Caus-Vid also leverages DMD~\cite{yin2024one} for 4-step video generation. However, the qualitative results indicate that our GAN object training scheme provides better initialization compared to the regression training scheme of ODE pairs. 
\vspace{-1em}
\paragraph{Quantitative Analysis}
\vspace{-1em}
\begin{table}[h]
    \small
    \centering
    \renewcommand{\arraystretch}{1.0}
    \setlength{\tabcolsep}{4pt}
    \begin{tabular}{clcccc}
        \toprule
        % & \multicolumn{4}{c}{\textbf{ArtifactInjection}} \\
        % \cmidrule(lr){2-5} 
        & & PSNR $\uparrow$ & SSIM $\uparrow$ & LPIPS $\downarrow$ & FID $\downarrow$ \\
        \midrule
        \multirow{3}{*}{\rotatebox[origin=c]{90}{\small{mip-360}}}
        &VC+DMD2 & 9.29 & 0.184 & 0.836 & 35.59  \\
        &VC+CausVid & 15.77 & 0.336 & 0.441 & 19.27  \\
        &Ours & \textbf{16.28} & \textbf{0.352} & \textbf{0.429} & \textbf{17.44}  \\

        \midrule
        \multirow{3}{*}{\rotatebox[origin=c]{90}{\small{TNT}}}
        &VC+DMD2 & 10.27 & 0.209 & 0.731 & 37.67 \\
        &VC+CausVid & 17.33 & 0.405 & 0.232 & 24.92  \\
        &Ours & \textbf{18.72} & \textbf{0.411} & \textbf{0.210} & \textbf{23.64} \\

        \bottomrule
    \end{tabular}
    \caption{\textbf{Quantitative Results.} We report four quantitative metrics for image quality comparison on 2 separate datasets: MipNeRF360~\cite{Barron2021Mip-NeRF:Fields} (denoted as mip-360 in the table) and Tanks-and-Temples~\cite{knapitsch2017tanks} (denoted as TNT in the table). \textbf{VC} stands for ViewCrafter~\cite{yu2024viewcrafter} for space saving purpose. \textbf{VC+X} means applying diffusion acceleration method X on ViewCrafter.}
    \label{tab:quant_vid}
\end{table}
Table~\ref{tab:quant_vid} presents quantitative comparisons with other diffusion distillation methods, and further proves the superiority of our method. Our method differs with DMD2~\cite{yin2025improved} in several ways: 1) we apply a 3D discriminator on video data, 2) we perform soften reverse-KL divergence for DMD optimization, 3) our GAN training and DMD training are kept separate for more stable training and more accurate density ratio calculation. Through training DMD2~\cite{yin2025improved}, we found a unstable training process and we recorded the last model before the training collapsed. In comparison, CausVid~\cite{yin2024slow} presents more stable training than DMD2~\cite{yin2025improved} but still suffer from mode collapse. Our method surpasses the other two baseline methods in both perceptual and distribution metrics.

\subsection{Ablation Studies}
% \vspace{-1em}
\begin{table}[h]
    \small
    \centering
    \renewcommand{\arraystretch}{1.0}
    \setlength{\tabcolsep}{2pt}
    \begin{tabular}{ccc|cccc}
        \toprule
        % & \multicolumn{4}{c}{\textbf{ArtifactInjection}} \\
        % \cmidrule(lr){2-5} 
        GAN& DMD & Soften RKL & PSNR $\uparrow$ &  SSIM $\uparrow$ & LPIPS $\downarrow$ & FID $\downarrow$ \\
        \midrule
        
        \xmark & \cmark & \cmark & 8.62 & 0.154 & 0.880 & 40.17  \\
        \cmark & \xmark & \xmark & 16.23 & 0.369 & 0.375 & 21.48 \\
        
        \cmark & \cmark & \xmark & 16.85 & \textbf{0.385} & 0.337 & 21.05  \\
        \cmark & \cmark & \cmark & \textbf{17.50} & 0.382 & \textbf{0.320} & \textbf{20.54}  \\

        \bottomrule
    \end{tabular}
    \caption{\textbf{Ablation studies.} Quantitative analysis of different components of FVGen. }
    \label{tab:ablation}
\end{table}
\begin{figure}
    \centering
    \includegraphics[width=\linewidth]{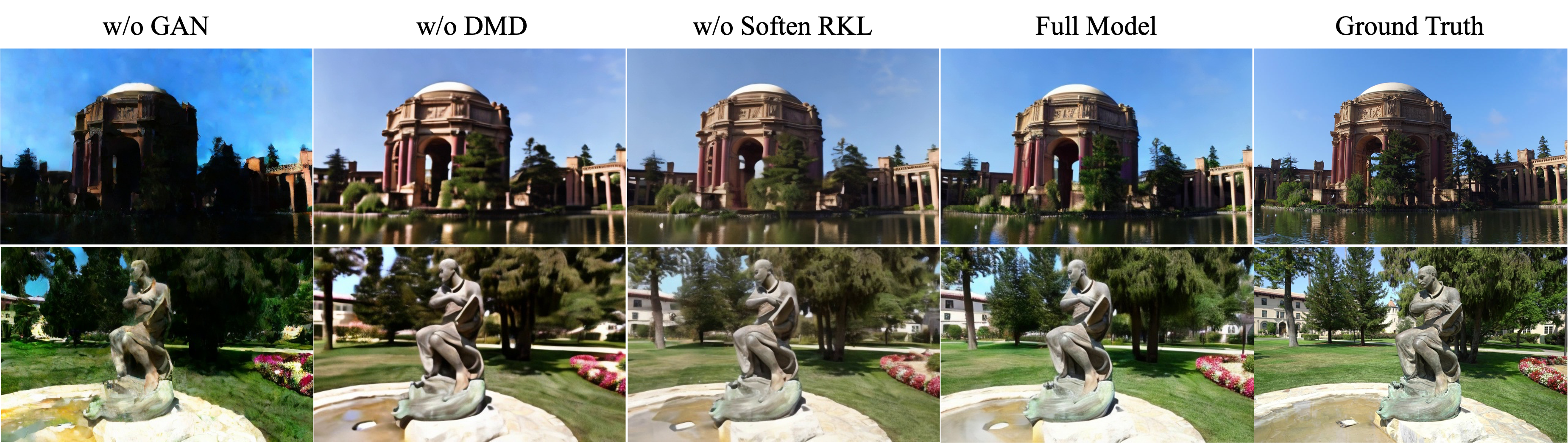}
    \caption{\textbf{Ablation Studies.} Visualization of the contribution of different components of FVGen. Our full model presents the most visual similarity between the ground truth. }
    \label{fig:ablation}
    \vspace{-1em}
\end{figure}
In this section, we analyze the contributions of each module of our proposed method in Table~\ref{tab:ablation} and Figure~\ref{fig:ablation}.
\vspace{-1em}
\paragraph{Assessing student initialization.} As it can be seen from the first column of Figure~\ref{fig:ablation} and Table~\ref{tab:ablation}, student initialization is a pivotal module of our architecture. Without student initialization with GAN optimization, the student model would fail to generate samples close to the teacher distribution, thus the teacher score function is not reliable for distillation.
\vspace{-1em}
\paragraph{Assessing distribution matching.} Optimization of DMD continues to align the student model distribution with teacher model. Comparing with full model, although the perceptual metrics are not significantly degraded, qualitative results show that without distribution matching, the generated results appear significant blurriness and artifacts.
\vspace{-2em}
\paragraph{Assessing soften reverse KL-divergence.} As discussed in Section~\ref{sec:f-divergence}, soften reverse KL-divergence enables more stable and robust training. Therefore, the DMD optimization achieves better convergence and generated results are more distilled into teacher distribution.

% \section{Conclusion and Limitations}
% In summary, FVGen is a novel framework to accelerate a state-of-the-art diffusion-based novel view synthesis model for few-step video frames generation. Our method leverages training a GAN objective for effective student model initialization and incorporates a soften reverse KL-divergence for a stable distribution matching between student and teacher model. While maintaining similar or slightly better visual quality, our method successfully decreases the sampling time by more than 90\%. However, we still identify several limitations in our current work. First, although FVGen achieves similar visual quality compared to ViewCrafter, it is limited by the drawbacks of ViewCrafter: e.g, degradation in structural integrity and consistency when experiencing extremely sparse inputs. Second, FVGen benefits from the synergy of three video diffusion models; therefore, due to computational constraints, we were only able to train the model with videos up to only 16 frames, which may not provide enough coverage for extremely large scenes. Therefore, our future work would address on the limitations mentioned above.

\section{Limitations}
% In summary, FVGen is a novel framework to accelerate a state-of-the-art diffusion-based novel view synthesis model for few-step video frames generation. Our method leverages training a GAN objective for effective student model initialization and incorporates a soften reverse KL-divergence for a stable distribution matching between student and teacher model. While maintaining similar or slightly better visual quality, our method successfully decreases the sampling time by more than 90\%. However, 

We identify several limitations in our current work. First, although FVGen achieves similar visual quality compared to ViewCrafter, it is limited by the drawbacks of ViewCrafter: e.g, degradation in structural integrity and consistency when experiencing extremely sparse inputs. Second, FVGen benefits from the synergy of three video diffusion models; therefore, due to computational constraints, we were only able to train the model with videos up to only 16 frames, which may not provide enough coverage for extremely large scenes. Therefore, our future work would address on the limitations mentioned above.
\section{Acknowledgement}
Supported by the Intelligence Advanced Research Projects Activity (IARPA) via Department of Interior/ Interior Business Center (DOI/IBC) contract number 140D0423C0075. The U.S. Government is authorized to reproduce and distribute reprints for Governmental purposes notwithstanding any copyright annotation thereon. Disclaimer: The views and conclusions contained herein are those of the authors and should not be interpreted as necessarily
representing the official policies or endorsements, either expressed or implied, of IARPA, DOI/IBC, or the U.S. Government.
{
    \small
    \bibliographystyle{ieeenat_fullname}
    \bibliography{main}
}
% ICCV 2025 Paper Template

\clearpage
\setcounter{page}{1}
\maketitlesupplementary

\noindent In this supplementary material, we mainly present the mathematical proof of Equation (8) in our main paper.
\newline\newline
\noindent Starting from Equation (6):

\begin{align}
    \mathbf{D}_\texttt{Soften-RKL}(p_{\text{fake},t}\| p_{\text{real},t})
    =\mathbf{D}_{KL}\bigg(\frac{1}{2}p_{\text{real},t}+\frac{1}{2}p_{\text{fake},t}\bigg\|p_{\text{real},t}\bigg). \nonumber
\end{align}

\noindent For notation simplicity purpose, we define $p_t(\boldsymbol{x}):=p_{\text{real},t}$ and $q_t(\boldsymbol{x}):=p_{\text{fake},t}$; here $\boldsymbol{x}=F(G_\theta(z),t)$ defined in the main paper. Then, we have:

\begin{align}
    &\mathbf{D}_\texttt{Soften-RKL}(p_t(\boldsymbol{x})\|q_t(\boldsymbol{x})) \nonumber \\
    &=\mathbf{D}_{KL}\bigg(\frac{1}{2}p_t(\boldsymbol{x})+\frac{1}{2}q_t(\boldsymbol{x})\|p_t(\boldsymbol{x})\bigg) \nonumber \\
    &=\int q_t(\boldsymbol{x})\bigg(\frac{p_t(\boldsymbol{x})}{q_t(\boldsymbol{x})}+1\bigg)\log\bigg(\frac{1}{2}+\frac{q_t(\boldsymbol{x})}{2p_t(\boldsymbol{x})}\bigg)d\boldsymbol{x}. \nonumber
\end{align}

\noindent Taking derivative w.r.t to model parameter $\theta$, we have the soften reverse KL divergence loss:

\begin{align}
&\nabla_\theta\mathcal{L}_{\text{DMD}}^{\texttt{Soften-RKL}} \nonumber \\
&=\nabla_\theta\int q_t(\boldsymbol{x})\bigg(\frac{p_t(\boldsymbol{x})}{q_t(\boldsymbol{x})}+1\bigg)\log\bigg(\frac{1}{2}+\frac{q_t(\boldsymbol{x})}{2p_t(\boldsymbol{x})}\bigg)d\boldsymbol{x} \nonumber \\
&=\int \nabla_\theta \bigg[q_t(\boldsymbol{x})\bigg(\frac{p_t(\boldsymbol{x})}{q_t(\boldsymbol{x})}+1\bigg)\log\bigg(\frac{1}{2}+\frac{q_t(\boldsymbol{x})}{2p_t(\boldsymbol{x})}\bigg)\bigg]d\boldsymbol{x} \nonumber \\
&=\underbrace{\int \nabla_\theta q_t(\boldsymbol{x})\bigg[\bigg(\frac{p_t(\boldsymbol{x})}{q_t(\boldsymbol{x})}+1\bigg)\log\bigg(\frac{1}{2}+\frac{q_t(\boldsymbol{x})}{2p_t(\boldsymbol{x})}\bigg)\bigg]d\boldsymbol{x}}_{A}  \nonumber \\
&+\underbrace{\int  q_t(\boldsymbol{x})\nabla_\theta\bigg[\bigg(\frac{p_t(\boldsymbol{x})}{q_t(\boldsymbol{x})}+1\bigg)\log\bigg(\frac{1}{2}+\frac{q_t(\boldsymbol{x})}{2p_t(\boldsymbol{x})}\bigg)\bigg]d\boldsymbol{x}}_{B}. \nonumber
\end{align}

\noindent As $p_t(\boldsymbol{x})$ is the real distribution, which is constant w.r.t $\theta$. According to the chain rule, we have:
\begin{align}
    B&=\int  q_t(\boldsymbol{x})\bigg[-\frac{p_t(\boldsymbol{x})}{q_t^2(\boldsymbol{x})}\log\bigg(\frac{1}{2}+\frac{q_t(\boldsymbol{x})}{2p_t(\boldsymbol{x})}\bigg) \nonumber \\
    &+\bigg(\frac{p_t(\boldsymbol{x})}{q_t(\boldsymbol{x})}+1\bigg)\frac{1}{p_t(\boldsymbol{x})+q_t(\boldsymbol{x})}\bigg]\nabla_\theta q_t(\boldsymbol{x})d\boldsymbol{x} \nonumber \\
    &=\int\nabla_\theta q_t(\boldsymbol{x})\bigg[-\frac{p_t(\boldsymbol{x})}{q_t(\boldsymbol{x})}\log\bigg(\frac{1}{2}+\frac{q_t(\boldsymbol{x})}{2p_t(\boldsymbol{x})}\bigg)+1\bigg]. \nonumber
\end{align}

\noindent Define $r_t(\boldsymbol{x})={p_t(\boldsymbol{x})}/{q_t(\boldsymbol{x})}$. Combining the derived $B$ and previous $A$, we have:

\begin{align}
    &\nabla_\theta\mathcal{L}_{\text{DMD}}^{\texttt{Soften-RKL}} \nonumber\\
    &=\int\nabla_\theta q_t(\boldsymbol{x})\bigg[\log\bigg(\frac{1}{2}+\frac{1}{2r_t(\boldsymbol{x})}\bigg)+1\bigg] d\boldsymbol{x} \nonumber \\
    &=\int q_t(\boldsymbol{x})\frac{2r_t(\boldsymbol{x})}{r_t(\boldsymbol{x})+1}\bigg(-\frac{1}{2r^2_t(\boldsymbol{x})}\bigg)\frac{\partial r_t(\boldsymbol{x})}{\partial\boldsymbol{x}}\frac{\partial G_\theta(\boldsymbol{z})}{\partial\theta}d\boldsymbol{x} \nonumber \\
    &=-\int q_t(\boldsymbol{x})\frac{1}{r_t(\boldsymbol{x})(r_t(\boldsymbol{x})+1)}\frac{\partial r_t(\boldsymbol{x})}{\partial\boldsymbol{x}}\frac{\partial G_\theta(\boldsymbol{z})}{\partial\theta}d\boldsymbol{x}. \nonumber
\end{align}

\noindent According to log-derivative trick, we have:
\begin{align}
\frac{\partial r_t(\boldsymbol{x})}{\partial \boldsymbol{x}}
&= \frac{%
      \displaystyle
      \frac{\partial p_t(\boldsymbol{x})}{\partial \boldsymbol{x}}\;q_t(\boldsymbol{x})
      \;-\;\frac{\partial q_t(\boldsymbol{x})}{\partial \boldsymbol{x}}\;p_t(\boldsymbol{x})
    }{%
      \bigl(q_t(\boldsymbol{x})\bigr)^{2}
    }
    \nonumber\\
&= \frac{%
      \displaystyle
      \frac{\nabla_{\boldsymbol{x}}\,p_t(\boldsymbol{x})}{p_t(\boldsymbol{x})}
      \;-\;\frac{\nabla_{\boldsymbol{x}}\,q_t(\boldsymbol{x})}{q_t(\boldsymbol{x})}
    }{%
      \displaystyle
      \frac{q_t(\boldsymbol{x})}{p_t(\boldsymbol{x})}
    }
    \nonumber\\
&= r_t(\boldsymbol{x})
   \bigl(\,
     \nabla_{\boldsymbol{x}}\log p_t(\boldsymbol{x})
     \;-\;\nabla_{\boldsymbol{x}}\log q_t(\boldsymbol{x})
   \bigr)
   \nonumber\\
&= r_t(\boldsymbol{x})\,
   \bigl[s_{\mathrm{real},t}(\boldsymbol{x})
     \;-\;s_{\mathrm{fake},t}(\boldsymbol{x})
   \bigr]. \nonumber
\end{align}

\noindent Therefore:
\begin{align}
    &\nabla_\theta\mathcal{L}_{\text{DMD}}^{\texttt{Soften-RKL}} \nonumber \\
    &=-\int q_t(\boldsymbol{x})\frac{1}{r_t(\boldsymbol{x})+1}\bigl[s_{\mathrm{real},t}(\boldsymbol{x})
     \;-\;s_{\mathrm{fake},t}(\boldsymbol{x})
   \bigr]\frac{\partial G_\theta(\boldsymbol{z})}{\partial\theta}d\boldsymbol{x} \nonumber \\
   &=-\mathbb{E}\bigg[\frac{1}{r_t(\boldsymbol{x})+1}\bigl(s_{\mathrm{real},t}(\boldsymbol{x})
     \;-\;s_{\mathrm{fake},t}(\boldsymbol{x})
   \bigr)\frac{\partial G_\theta(\boldsymbol{z})}{\partial\theta}\bigg] \nonumber
\end{align}

\noindent This concludes the proof of Equation (8) in the main paper.

\end{document}